\documentclass[10pt,twocolumn,letterpaper]{article}

\usepackage{cvpr}
\usepackage{times}
\usepackage{epsfig}
\usepackage{graphicx}
\usepackage{amsmath}
\usepackage{amssymb}

\usepackage{multirow}
\usepackage{booktabs} % for pretty plots

\usepackage[breaklinks=true,bookmarks=false]{hyperref}

\cvprfinalcopy % *** Uncomment this line for the final submission

\def\cvprPaperID{****} % *** Enter the CVPR Paper ID here

\begin{document}

%%%%%%%%% TITLE
\title{Collaborative Learning for Faster StyleGAN Embedding}

\author{Shanyan Guan\\
% Shanghai Jiao Tong University\\
{\tt\small shyanguan@sjtu.edu.cn}
\and
Ying Tai\\
% Youtu Lab, Tencent\\
{\tt\small yingtai@tencent.com}
\and
Bingbing Ni\\
% Shanghai Jiao Tong University\\
{\tt\small nibingbing@sjtu.edu.cn}
\and
Feida Zhu\\
% Youtu Lab, Tencent\\
{\tt\small zhufeida@connect.hku.hk}
\and
Feiyue Huang\\
% Youtu Lab, Tencent\\
{\tt\small garyhuang@tencent.com}
\and
Xiaokang Yang\\
% Shanghai Jiao Tong University\\
{\tt\small xkyang@sjtu.edu.cn}
}

\maketitle
%\thispagestyle{empty}

%%%%%%%%% ABSTRACT
\begin{abstract}
   The latent code of the recent popular model StyleGAN has learned disentangled representations thanks to the multi-layer style-based generator. Embedding a given image back to the latent space of StyleGAN enables wide interesting semantic image editing applications. Although previous works are able to yield impressive inversion results based on an optimization framework, which however suffers from the efficiency issue. In this work, we propose a novel collaborative learning framework that consists of an efficient embedding network and an optimization-based iterator.  On one hand, with the progress of training, the embedding network gives a reasonable latent code initialization for the iterator. On the other hand, the updated latent code from the iterator in turn supervises the embedding network. In the end, high-quality latent code can be obtained efficiently with a single forward pass through our embedding network. Extensive experiments demonstrate the effectiveness and efficiency of our work.
\end{abstract}

\section{Introduction}
Generative Adversarial Networks~\cite{goodfellow2014generative} has been widely applied in various image processing tasks to synthesize realistic images, such as image-to-image~\cite{isola2017image,zhu2017toward,zhu2017toward,wang2018high,liu2017unsupervised}, and semantic attribute editting~\cite{choi2018stargan,li2018beautygan,shen2019interpreting,qian2019make,liu2019stgan}. With the rapid progress of high-quality image generative models \cite{karras2017progressive,karras2019style,karras2019analyzing}, reusing a well-trained model as tools for image manipulation has attracted more attention in the computer vision community. Particularly, inspired by Adaptive Instance Normalization (AdaIN) \cite{huang2017arbitrary}, the StyleGAN~\cite{karras2019style} exposes multi-layer intermediate latent codes to control the image synthesis process. The intermediate latent space is shown to contain disentangled semantics~\cite{karras2019style}. As a result, once inverting the given image to the latent code of StyleGAN, we can make the semantic modification to the given image by editing the corresponding latent code.

There have been several optimization-based approaches~\cite{abdal2019image2stylegan++,abdal2019image2stylegan,git2} that attempt to embed a given image into the StyleGAN latent space. Specifically, they start from an initialized latent code, and then optimize the latent code to minimize the difference between the input image and the synthesized image through error back-propagation. Although they got reasonable embeddings of the input images, there still exist two main drawbacks: (1) The optimization procedure is time-consuming, which at least takes several minutes on a modern GPU. (2) The final result is sensitive to the choice of the initialization.

\begin{figure*}[t]
\centering
\includegraphics[width=\linewidth]{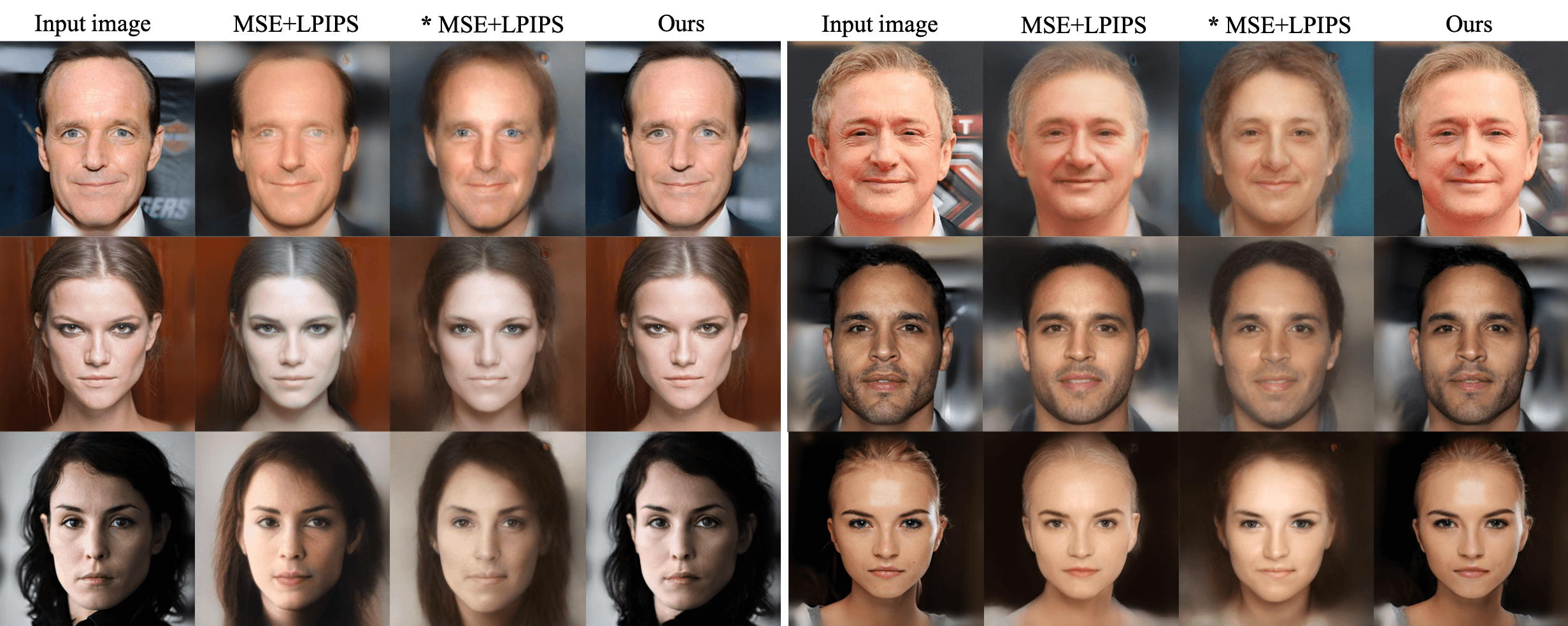}
\caption{Evalution on the effect of latent-level loss $\mathcal{L}_{w}$ (Eq.~\ref{eq:wloss}). The results demonstrate that only use image/feature-level losses, without the supervision on the latent code, is not enough to accurately inverts images into the latent space of StyleGAN, whether the generator is trained together or not. We set three baselines illustrated as follows. MSE: training the embedding network with MSE loss. MSE + LPIPS: training the embedding network with MSE loss and LPIPS~\cite{zhang2018unreasonable}. *MSE + LPIPS: training the generator of StyleGAN together with the embedding network using MSE loss and LPIPS.}
\label{fig:loss-gen-effect}
\end{figure*}

Alternatively, we shift our efforts towards training an embedding network to learn the inverse mapping from the image space to the latent space. Once trained, the embedding can be done in real-time, without any initialization need on the latent code. However, training such an embedding network is not trivial since it should be able to infer reasonable latent codes for a wide range of images. Besides, the conventional image/feature-level constraints (\ie MSE loss and Perceptual loss ~\cite{johnson2016perceptual}) between the input image and the reconstructed images from the embedded latent codes are not strong enough to guide the embedding network. As shown in Fig.~\ref{fig:loss-gen-effect}, the reconstructed images vary obviously with the input images. We also have tried to finetune the synthesis network of StyleGAN along with training, which turned out to be a useless attempt.

In this paper, we propose a novel collaborative learning framework for efficient image embedding. The framework consists of an embedding network and an optimization-based iterator. These two components cooperate tightly to form one training loop. Given one training sample, the embedding network firstly infers its latent code, which is further used to initialize the iterator. Then, the iterator optimizes the latent code to an optimum. The updated latent code, together with the image/feature-level losses, is utilized to supervise the embedding network. With the progress of training, the embedding network learns to generate more accurate latent code, which also accelerates the optimization steps inside the iterator. As shown in Fig.~\ref{fig:loss-gen-effect}, our method greatly improves the quality of the embedded latent codes.

Moreover, we propose a new embedding network structure. We design two separate encoders inside the embedding network to encode the face identities and face attributes information, respectively. The identity features and attribute features are merged carefully through denormalization operation. Finally, a regressor is used to map the merged features to the latent code.

We summarize the contribution of our approach as follows:

(1) We propose a novel collaborative learning framework to train our embedding network in unsupervised case.

(2) The carefully-designed embedding network is able to map real images into the latent space of StyleGAN effectively and efficiently. 

(3) With similar performance. our model is about 500 times faster than the current efficient model. Moreover, broad semantic manipulation applications have been explored to demonstrate the potential of our approach.

\section{Related Work}
\noindent \paragraph{\textbf{Embedding of Generative Models.}} Generative models usually use adversarial training to generate high-resolution images from latent codes \cite{radford2015unsupervised,karras2017progressive,miyato2018spectral,brock2018large}. The latent space may exhibit meaning properties that control the attributes of generated data. However, the ability to find an effective latent code that reconstructs a given image is not ensured for GANs \cite{creswell2018inverting}.  Recently, optimization-based methods, \eg Image2StyleGAN and Image2StyleGAN++\cite{abdal2019image2stylegan,abdal2019image2stylegan++} successfully embedded images into the latent space of StyleGAN \cite{karras2019style} and showcased interesting applications by manipulating the latent code. Besides, Karras et al. introduced StyleGAN2 \cite{karras2019analyzing} which further improved the quality of the matching latent code. However, these optimization-based methods share the same drawback of high computation complexity, which takes several minutes on a modern GPU. In contrast, our embedding network only takes less than 1 second in a single forward pass, which is about 500 times faster.

\noindent \paragraph{\textbf{Collaborative Learning Works.}} The main idea of Collaborative Learning is to build share learning branches to obtain more informative features. Previous works can be split into two categories. One is to simultaneously train multi-head models and then merge their features to obtain more useful features~\cite{song2018collaborative,sheng2019unsupervised}. Another is to unsupervised train a model by excavate the same concepts from different environment, which has been used in various task, \eg image classification~\cite{chen2020simple,batra2017cooperative,zhou2019collaborative}, unsupervised domain adaptation~\cite{zhang2018collaborative}, subspace clustering~\cite{zhang2019neural} and reinforcement learning~\cite{kong2017collaborative}. However previous works are mainly designed to learn semantic features but fail to be applied in the image inverting task since it explicitly requires pixel-level inverting. Instead, our collaborative learning framework first to combine the deep learning model with the optimization-based approach to realize the non-trivial StyleGAN inverting task with pixel-level accuracy.

\begin{figure*}[t]
\centering
\includegraphics[width=0.8\linewidth]{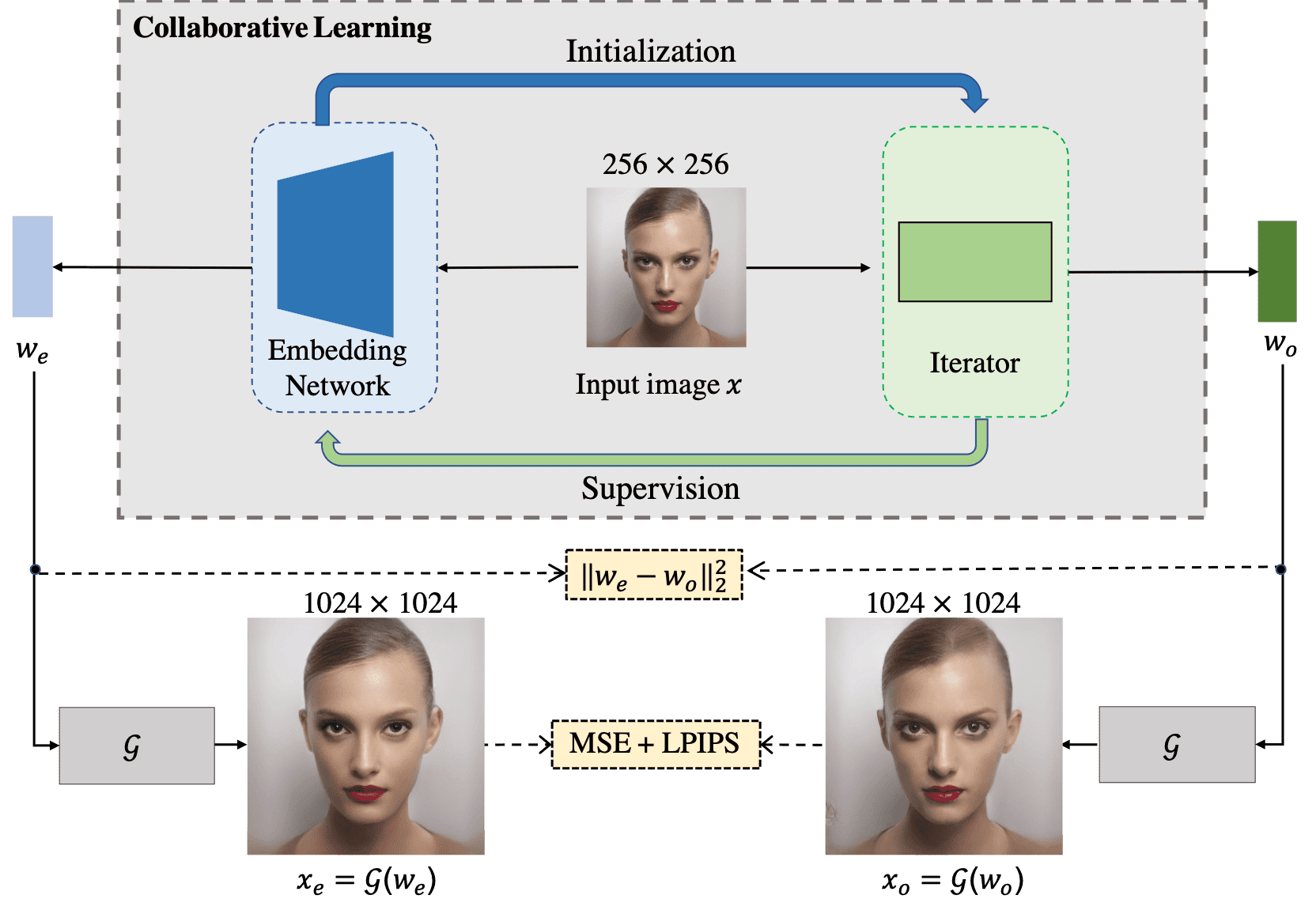}
\caption{The overview of our colleborative learning framework. Given an input image $x$ in $256 \times 256$ resolution, the embedding network generates its latent code $w_e$, which is then send to as the initialization of the iterator. The output of iterator $w_o$ in turns supervise the training of the embedding network, using MSE loss, LPIPS loss and latent code loss $||w_e-w_o||_2^2$.} 
\label{fig:mainfig}
\end{figure*}

\section{Methodology}
Given a real image, our goal is to efficiently learn its latent code in StyleGAN latent space, from which we can realize various semantic image modification in real-time. Considering the consensus in previous works~\cite{abdal2019image2stylegan,git2,abdal2019image2stylegan++}, we chose the $W+ \in \mathbb{R}^{18\times512}$ space~\cite{abdal2019image2stylegan} as the target latent space. 
To achieve this goal, we propose a collaborative learning framework shown in Fig.~\ref{fig:mainfig}, which consists of an embedding network and an optimization-based iterator. Given an image, the embedding network generates its latent code, which is then sent to initialize the iterator. After iterative optimization, the final output of iterator in turn feedbacks to the embedding network as supervision. 

In the following, we first introduce the iterator (Sec.~\ref{sec:sec_iterator}), and then describe embedding network (Sec.~\ref{sec:sec_network}). Finally, we illustrate the details of our collaborative learning framework and discuss its characteristics (Sec.~\ref{sec:sec_clf}). 

\begin{figure}[t]
\centering
\includegraphics[width=0.95\linewidth]{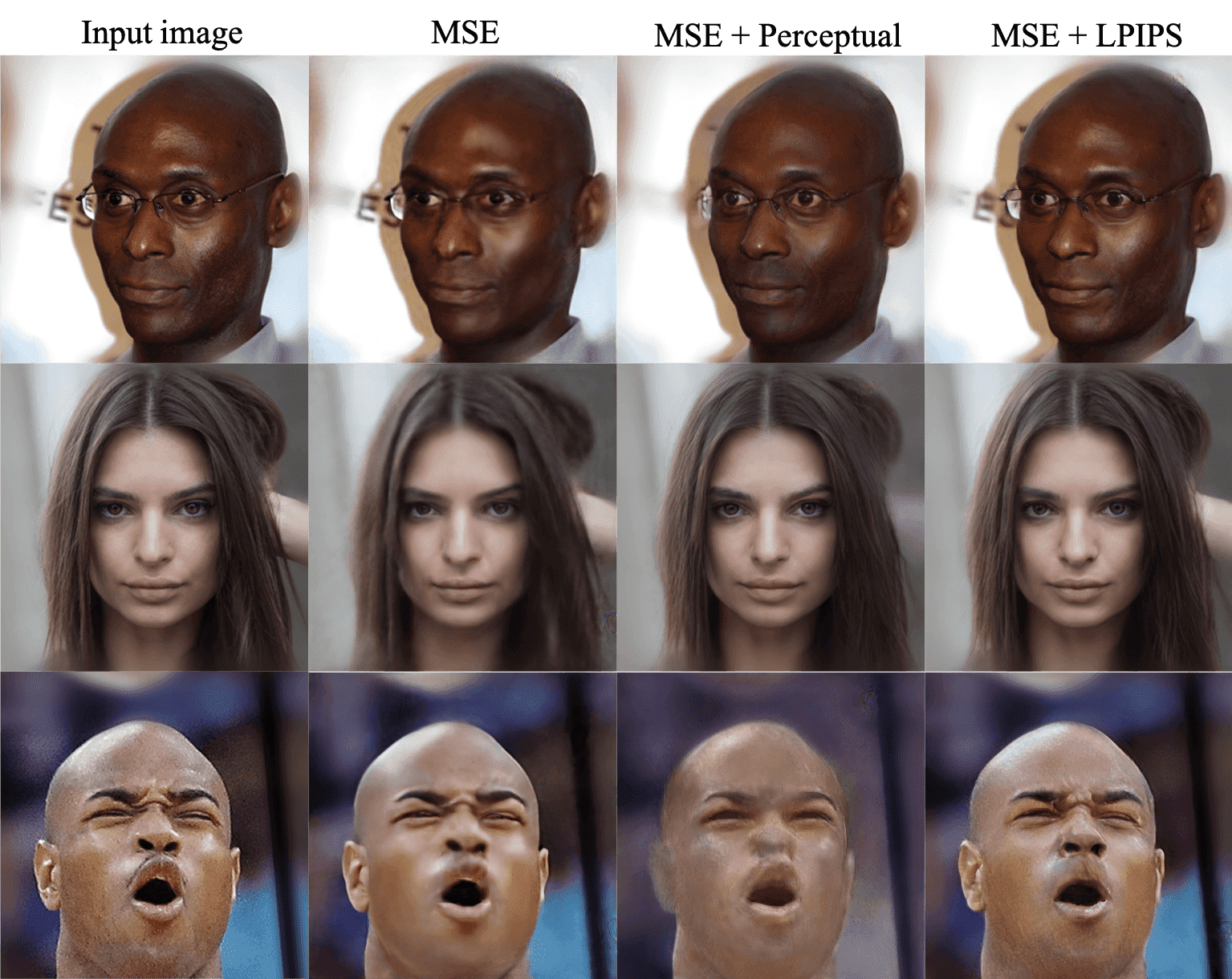}
\caption{Visual comparison against the effects of losses in the iterator. We can clearly observe that only using MSE loss, the inverted images are blurred. After adding Perceptual loss, the inverted results are more clear, but the artifacts still exist, \eg the glasses in the first column. Replacing the perceptual loss with LPIPS loss, the inverted image is more clear and complete than using Perceptual and MSE losses.}
\label{fig:per-vs-lpips}
\end{figure}

\subsection{Iteraor: Optimization-Based Embedding Approach}
\label{sec:sec_iterator}
The iterator in our collaborative learning framework has the same design with Image2StyleGAN but has two differences: (1) the initialization comes from the embedding network, instead of a mean latent code. (2) We replace the Perceptual loss with LPIPS loss~\cite{zhang2018unreasonable} since we empirically observed better effect (as shown in Fig.~\ref{fig:per-vs-lpips}). 
Specifically, starting from the initialization, the iterator searches for the optimal latent code $w_o$ by minimizing the MSE and LPIPS losses between the given image $x$ and the generated image from the optimized latent code. The objective function for the iterator is:
\begin{align}
  \mathcal{L}_{opt} = ||\mathcal{G}(w) - x||_2^2 + \alpha \Phi(\mathcal{G}(w), x),
  \label{eq:opt}
\end{align}
where $w \in W+$ is the latent code to be optimized, $\mathcal{G}$ is a frozen generator of StyleGAN pretrained on FFHQ dataset~\cite{karras2019style}, $\Phi(\cdot)$ is the LPIPS loss~\cite{zhang2018unreasonable}, and the loss weight $\alpha$ is set as $1$.

\noindent \paragraph{\textbf{Weakness.}} Although the iterator can get a good fit by minimizing $\mathcal{L}_{opt}$, it's too slow to limit its wider applications in practice and is also sensitive to the initialization which leads to unstable performance. 

\begin{figure}[t]
\centering
\includegraphics[width=0.95\linewidth]{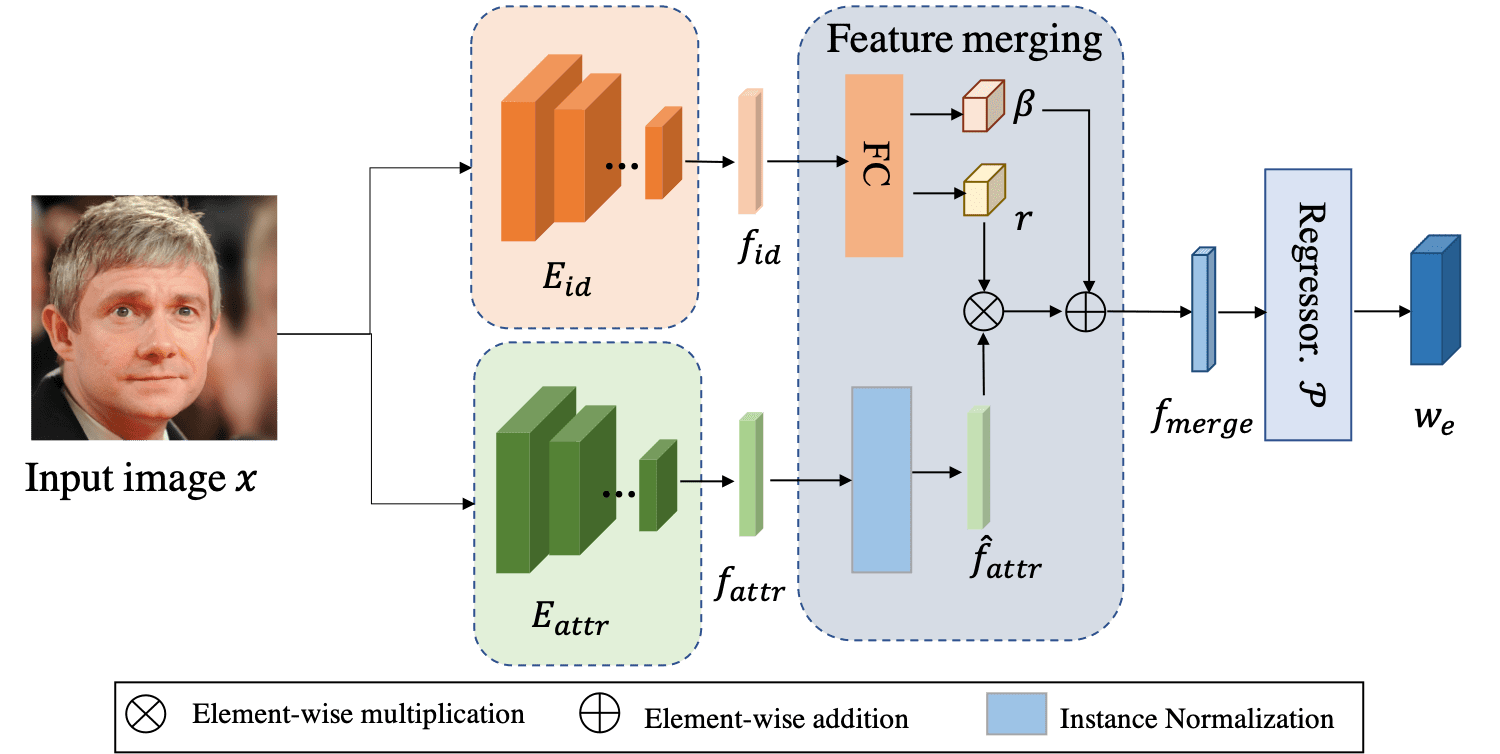}
\caption{Illustration of embedding network. Given an input image $x$, the embedding network learns to map it to a latent code $w_e$.}
\label{fig:structure-embeddingnetwork}
\end{figure}

\subsection{Embedding Network}
\label{sec:sec_network}
Fig.~\ref{fig:structure-embeddingnetwork} exhibits the structure of the embedding network. 
The embedding network mainly consists of three components: (1) the identity encoder $E_{id}$ to extract identity from the input image $x$, (2) the attribute Encoder $E_{attr}$ to extract attributes from the input image $x$ , and (3) the latent code regresser $\mathcal{P}$ to map extacted feature to a latent code $w_e \in W+$.

We use a pretrained Arcface model~\cite{deng2019arcface} except the final fully-connected layer as identity encoder, and the identity feature is defined as $f_{id} = E_{id}(x; \theta_{1})$, where $\theta_{1}$ denotes the parameters of $E_{id}$. Pretraining on large scale of face data , Arcface model can provide representative identity feature. The attribute encoder is the first five convolutional stages of ResNet-50~\cite{he2016deep}, and the attribute feature is written as: $f_{attr} = E_{attr}(x; \theta_{2})$, where the $\theta_{2}$ denotes the parameter of $E_{attr}$. Inspired from SPADE~\cite{park2019semantic} and StyleGAN~\cite{karras2019style}, we merge the identity feature and the attribute feature thourgh denormalization operation. After that the regressor $\mathcal{P}$, a tree-connected structure~\cite{richter2018treeconnect}, maps the merged feature $f_{merge}$ to the latent code $w_e$. Next we introduce the feature merging process.

Let the size of attribute feature $f_{attr}$ be $C \times H \times W$, where $C$ is the number of channels and $H \times W$ is the spatial dimention. We perform instance normalization~\cite{ulyanov2016instance} on $f_{attr}$:
\begin{align}
  \hat{f}_{attr} = \frac{f_{attr} - \mu}{\sigma}
\end{align}
where $\mu \in \mathbb{R}^{C}$ and $\sigma \in \mathbb{R}^{C}$ are the means and standard deviation of $f_{attr}$ along channel dimension. Then we intergrate the identity feature $f_{id}$ by denormalization, which is formulated as:
\begin{align}
  f_{merge} = \gamma * \hat{f}_{attr} + \beta
\end{align}
where $\gamma \in \mathbb{R}^{C}$ and $\beta \in \mathbb{R}^{C}$ are two modulation parameters generated from $f_{id}$ through a fully-connnection layer. 

\noindent \paragraph{\textbf{Weakness.}} Without the direct supervision of latent codes and using only the image/feature-level losses (\eg MSE and/or LPIPS loss), it is difficult for the embedding network to map the image to the latent space accurately. Since the AdaIN mechanism of StyleGAN takes statistics as input, the gradient from StyleGAN through image/feature-level losses can only let the embedding network map the approximate content of the image to the latent space, but not all the details, as shown in Fig.~\ref{fig:loss-gen-effect}. 

\subsection{Collaborative Learning Framework}
\label{sec:sec_clf}
Fig.~\ref{fig:mainfig} shows an overview of the collaborative learning framework. Given a real image $\boldsymbol{x}$, the embedding network maps it to the latent code $\boldsymbol{w}_e$. Then $\boldsymbol{w}_e$ is used to initialize the iterator. The iterator is time-consuming if it starts from a mean latent code but can be accelerated if initiated by a more proper latent code. This point is one of the foundations of our framework. Next, the optimized latent code $\boldsymbol{w}_o$ from the iterator in turn supervises the embedding network to produce more accurate latent codes. Two modules form a positive cycle to promote together and thus bypass the need for paired latent codes. The results of iterator supervise the embedding network on latent code level:
\begin{align}
  \mathcal{L}_{\boldsymbol{w}} = ||\boldsymbol{w}_e - \boldsymbol{w}_o||_2^2,
  \label{eq:wloss}
\end{align}
image level (\textit{i.e.}, MSE loss):
\begin{align}
  \mathcal{L}_{mse} &= ||\boldsymbol{x}_e - \boldsymbol{x}_o||_2^2,
\end{align}
and feature level (\ie, LPIPS loss):
\begin{align}
  \mathcal{L}_{per} &= \Phi(\boldsymbol{x}_e, \boldsymbol{x}_o)
\end{align}
where $\boldsymbol{x}_e = \mathcal{G}(\boldsymbol{w}_e)$ and $\boldsymbol{x}_o = \mathcal{G}(\boldsymbol{w}_o)$ are generated from $\boldsymbol{w}_e$ and $\boldsymbol{w}_o$ by the StyleGAN generator $\mathcal{G}$. 

\noindent \paragraph{\textbf{Learning.}} In summary, the total loss function of embedding network is:
\begin{align}
  \mathcal{L} = \lambda_1 \mathcal{L}_{mse} + \lambda_2 \mathcal{L}_{per} + \lambda_3 \mathcal{L}_{\boldsymbol{w}}
\end{align}
where  $\{\lambda_1, \lambda_2, \lambda_3\}$ are the loss weights. The loss function of the iterator is $\mathcal{L}_{opt}$ (Eq.~\ref{eq:opt}). We run the iterator 100 steps for each training batch. In addition, during the training, the iterator may produce worse optimization results than before. In order to ensure the embedding network with the most accurate supervision signal, we adopt a \textit{cache mechanism} to save the best results. If the current optimization is better than the cache, we take it as the new supervision, but if it is worse, we ignore it.

\noindent \paragraph{\textbf{Characteristics.}} % (fold)
\label{par:paragraph_name}
(1) In previous works~\cite{abdal2019image2stylegan,git2,abdal2019image2stylegan++}, the optimization is very slow since they use a mean latent code as initialization. In contrast, we provide a more reasonable latent code from the embedding network, which greatly speeds up the optimization. (2) In our framework, the embedding network cooperates tightly to form a self-improved training loop. That is, the more accurate $\boldsymbol{w}_o$ can supervise the embedding network to learn to embed images better, meanwhile the more reasonable $\boldsymbol{w}_e$ from the embedding network lead the iterator to produce $\boldsymbol{w}_o$ that is closer to the optimal. Detailed analysis and evaluation can be found in Sec. 4.2.

\noindent \paragraph{\textbf{Difference to the Off-line Pipeline.}}
The off-line pipeline, which uses the iterator to find out the latent representation of all images before training, which we regard to be inefficient and impractical for large-scale datasets. Refer to the reported time cost in Image2StyleGAN~\cite{abdal2019image2stylegan}, it's time-consuming for the off-line iterator to get satisfactory results when taking the mean latent code as initialization, and it gets lots of artifacts if the optimization steps are shortened. Instead, our online updating idea bypasses the dilemmas for the fact that we can dynamically provide better initializations to the iterator. 

\section{Experiments}
\noindent \paragraph{\textbf{Implementation.}} For each face image, we first crop and align the face following the the StyleGAN \cite{karras2019style} setting, and then resize them to $256 \times 256$. We implement our framework with PyTorch library~\cite{paszke2017automatic}. In all experiments, we use the Adam optimizer~\cite{Kingma2014Adam} with $lr=0.0001$ and $(\beta_1, \beta_2) = (0.5, 0.999)$. Please refer to the Supp. Mat. for more details about the network architectures and training procedures. 

We evaluate our framework on two datasets: CelebA-HQ \cite{karras2017progressive} and CACD \cite{chen2014cross}, which represent challenges in different aspects. CelebA-HQ is a high-definition dataset, which contains 30,000 images in $1024 \times 1024$ resolution. CACD has more than 160,000 images of low-quality images. For each dataset, 80\% images are randomly selected as the training set while the remaining images are used as the testing set. 

\noindent \paragraph{\textbf{Metrics.}} The emphasis of our framework is to accelerate the Image2StyleGAN embedding procedure meanwhile make the inverted image similar to the input image. Consequently, we comprehensively evaluate the effect of our framework in three aspects: (1) runtime, (2) pixel-level similarity measured through Peak Signal-to-Noise Ratio (PSNR) and SSIM, (3) perceptual-level similarity measured through LPIPS that is consistent with human perception.

\noindent \paragraph{\textbf{Evaluation Setup.}} We compare our method with StyleGAN-Encoder~\cite{git2}, Image2StyleGAN~\cite{abdal2019image2stylegan} and Image2StyleGAN++~\cite{abdal2019image2stylegan++}. Image2StyleGAN and StyleGAN-Encoder have the similar optimization framework. Image2StyleGAN takes the mean latent code as initialization while StyleGAN-Encoder takes the output of a pretrained ResNet-50 model~\cite{he2016deep}. Different from Image2StyleGAN and StyleGAN-Encoder, Image2StyleGAN++ not only optimizes the latent code but also optimizes the noise variables in the synthesis network of StyleGAN. Since Image2StyleGAN and Image2StyleGAN++ haven't published their codes, we implemented their methods with PyTorch. For a fair comparison, we exactly follow their experimental setup and do not change their training procedure. The codes are also included in the supplementary material for the check.

\subsection{Comparison against state-of-the-art Methods}

\begin{figure}[t]
\centering
\includegraphics[width=\linewidth]{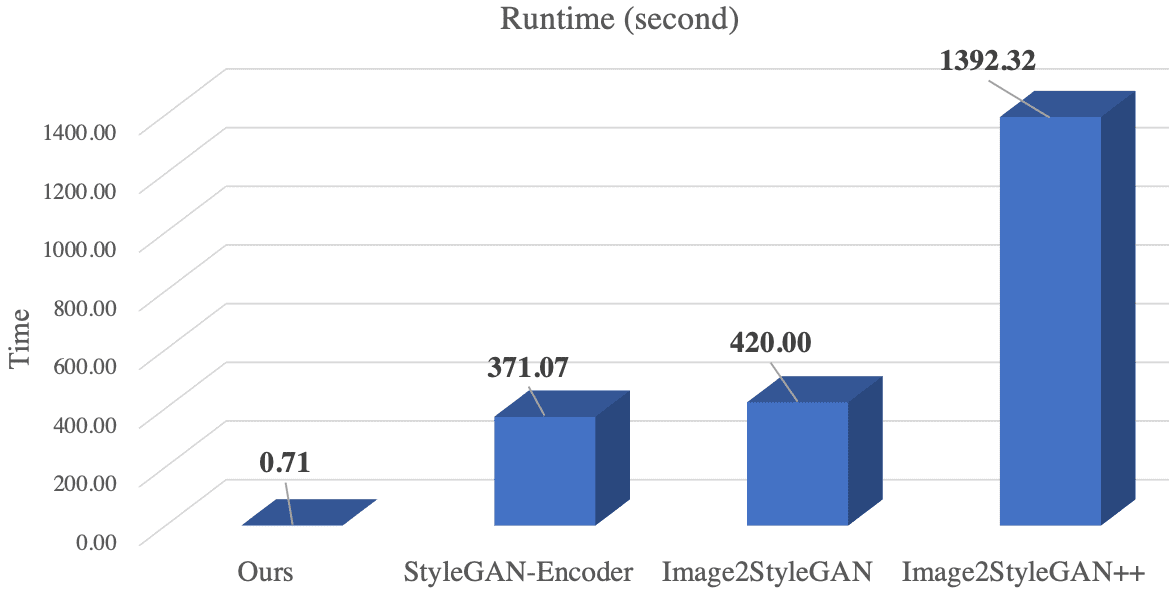}
\caption{Evaluation of the runtime (second). Compare with state-of-the-arts approaches, our method is the most efficient model, which infer the embedded latent code in less than one second. Besides, StyleGAN-Encoder is faster than Image2StyleGAN thanks to the better initialization provided from the pretrained ResNet-50. Keeping the same config in Image2StyleGAN, we statistic the runtime in a Tesla V100 GPU. Notice that the runtime of Image2StyleGAN here is the official report in their paper (about 7 minutes).} 
\label{fig:exp-runtime}
\end{figure}

In this section, our goal is to verify whether our model is faster than the state-of-the-art (SOTA) methods meanwhile achieves competitive embedding accuracy.

\noindent \paragraph{\textbf{Runtime:}}
For more convincing experimental results, the experiments are conducted on the same Tesla V100 GPU, and we directly cite the official runtime of Image2SytleGAN reported in their paper.
The runtime of our method and previous methods are reported in Fig.~\ref{fig:exp-runtime}, which clearly demonstrates the efficiency of our method. In contrast, optimization-based approaches are slow due to hundreds of optimization iterations. Image2StyleGAN++~\cite{abdal2019image2stylegan++} is even slower since it needs to optimize extra noise variables.

\begin{table}[t]
\centering
\label{tab:sim}
\resizebox{\linewidth}{!}{
    \begin{tabular}{@{}l@{\hspace{.2in}}c@{\hspace{.1in}}c@{\hspace{.1in}}c@{\hspace{.2in}}c@{\hspace{.1in}}c@{\hspace{.1in}}c@{}}
    \toprule
                        & \multicolumn{3}{c}{CelebA-HQ} & \multicolumn{3}{c}{CACD} \\
                        \cmidrule(r){2-4} \cmidrule(r){5-7} 
                        & PSNR (dB) ($\uparrow$) & SSIM ($\uparrow$) & LPIPS ($\downarrow$) & PSNR (dB) ($\uparrow$) & SSIM ($\uparrow$) & LPIPS ($\downarrow$) \\
    \midrule
    Image2StyleGAN         & 29.72 & 0.75 & 0.18  & 31.39 & 0.80 & 0.12  \\
    StyleGAN-Encoder     & 32.08 & 0.85 & 0.18  & 33.10 & 0.85 & \textbf{0.11} \\
    Image2StyleGAN++     & \textbf{32.46} & \textbf{0.90} & 0.22  & \textbf{34.40} & \textbf{0.90} & 0.15 \\
    Ours                 & 31.47 & 0.83 & \textbf{0.16}  & 32.05 & 0.83 & \textbf{0.11} \\
\bottomrule
\end{tabular}
}

\caption{Quantitative comparison of different embedding methods in terms of PSNR, SSIM, and LPIPS. The results indicate that our model achieves competitive performance. However, our model is about \textbf{500} times faster than the most efficient model.}
\end{table}

\begin{figure*}[t]
\centering
\includegraphics[width=0.9\linewidth]{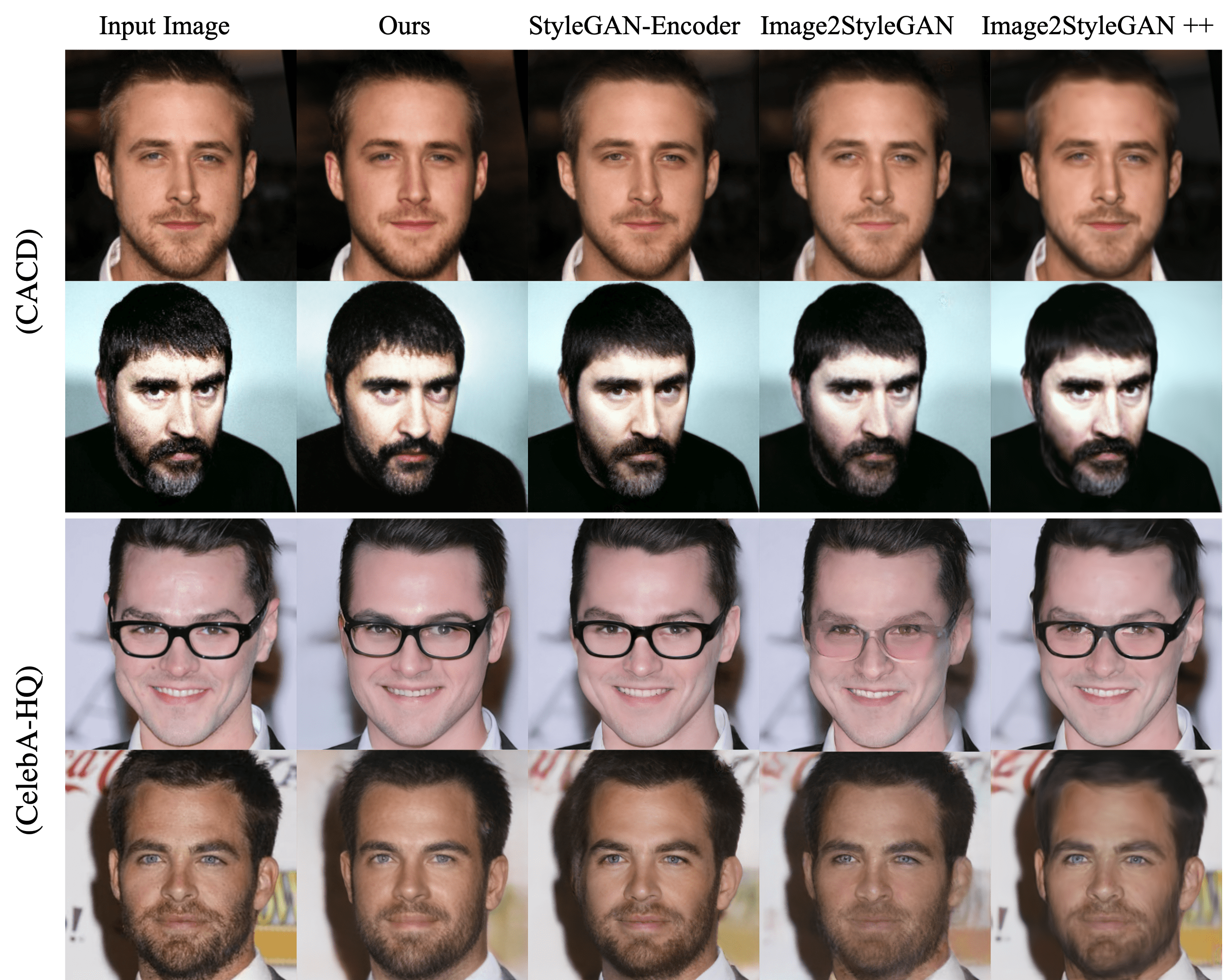}
\caption{Quantitative comparasion with baselines on CACD and CelebA-HQ datasets.}
\label{fig:mainfig}
\end{figure*}

\noindent \paragraph{\textbf{Quantitative Evaluation:}} Quantitative evaluation results are reported in Tab.~\ref{tab:sim}. For real images the have no paired latent codes, the most straightforward and fair scheme to evaluate the embedding accuracy is directly measuring the similarity between the input image and the generated image from the inverted latent code. Consequently, we comprehensively compare the inverting effect from pixel-level metric to perceptual metric, whose results are reported in Tab.~\ref{tab:sim}. From Tab.~\ref{tab:sim} we have serval observations:
\begin{enumerate}
    \item StyleGAN-Encoder performs better than Image2StyleGAN in all metrics. This is because StyleGAN-Encoder takes a customized initialization from a fine-tuned ResNet50 model for each image rather than a constant mean latent code. 
    \item The Image2StyleGAN++ achieves the best results in terms of PSNR and SSIM, but does not perform well in terms of LPIPS. We think the reason is that it only uses MSE loss but ignores to reduce the perceptual-aware error during noise optimization.
    \item Our method achieves the lowest LPIPS error and competitive performance in terms of PSNR and SSIM, compared with previous SOTA methods.
\end{enumerate}

\noindent \paragraph{\textbf{Qualitative Evaluation.}} A more detailed visual comparison between our method and previous methods is shown in Fig.~\ref{fig:mainfig}. Please refer to the supplemental material for more examples. The reconstructed images from the embedded latent codes, inferred by our method, are as close to the input images as the SOTA methods. It is worth to mention that our method infers the embedded latent code in less than one second, greatly faster than the previous methods. 

\subsection{Ablation Study}

\noindent \paragraph{\textbf{The Effect of Initialization on the Iterator.}} 
\label{subsec:init-ablates}
To examine that better initialization leads to faster convergency and better optimization results, we compare the effect of optimization using three different initialization schemes: random initialization (\ie Random), initialized by a mean latent code (\ie Mean), and initialized by the output by our embedding network (\ie Ours). We choose the optimization results of step 10, step 20, step 50 and step 100 to show the trend of the optimization procedure. Quantitative comparison results plotted in Fig.~\ref{fig:init-ablates-table} demonstrates that:
\begin{enumerate}
    \item Since the embedding network provides accurate initialization, the iterator fastly converges to the optimum at the early stage. This indicates that the better initialization makes the iterator more easily to find the optimal latent code.
    \item Analyzing the PSNR and SSIM, random initialization and average initialization have similar performance. However, recently several works~\cite{mathieu2015deep,denton2017unsupervised,lee2018stochastic,zhang2018unreasonable} argue that PSNR and SSIM are not convincing enough to measure the similarity between the input image and the generated image. To this end, we more concerned about the LPIPS metric. The average initialization achieves a higher LPIPS score which indicates its results are more acceptable for human perception.
\end{enumerate}

\begin{figure}[t]
\centering
\includegraphics[width=\linewidth]{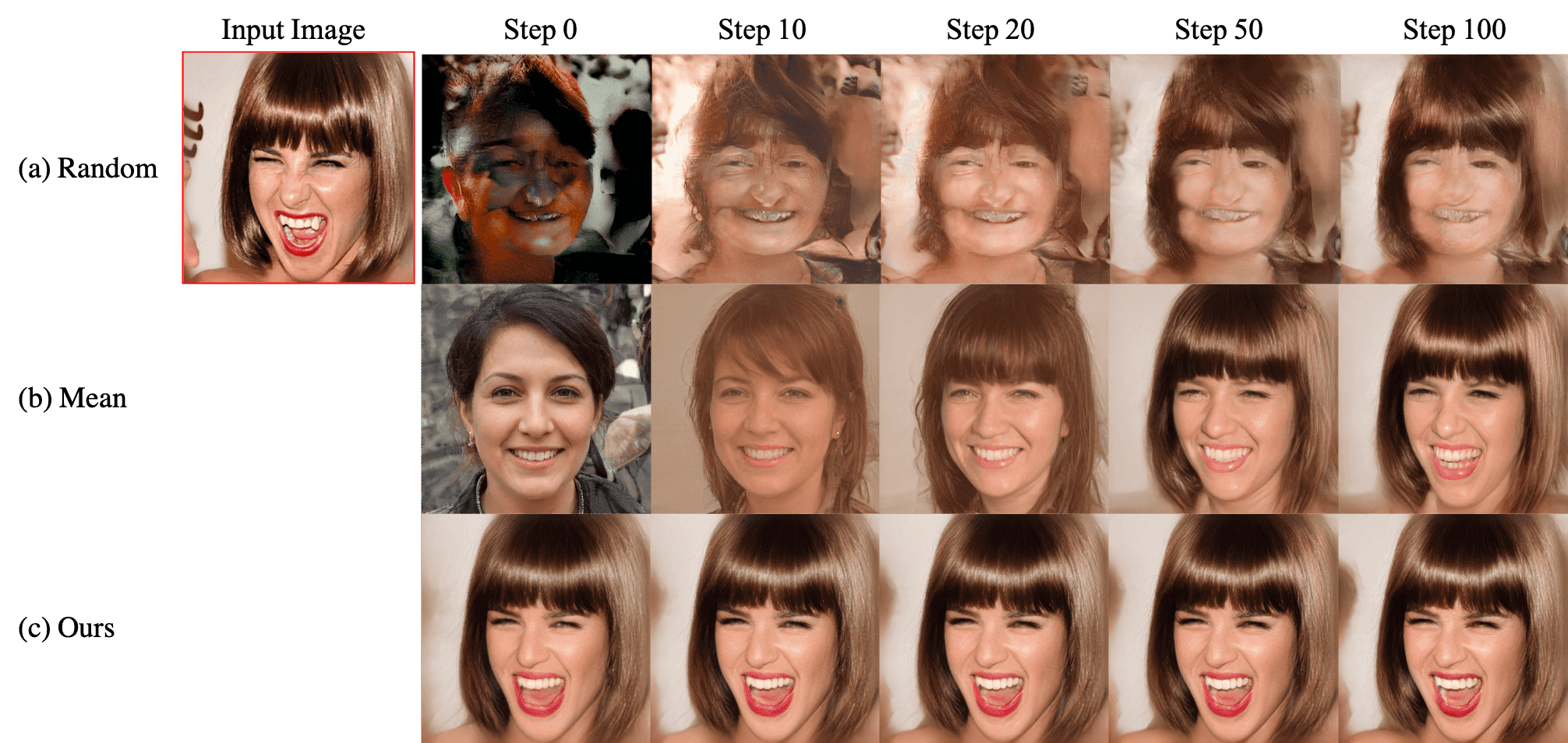}
\caption{Qualitative comparison with different intilization schme. The random initlization produce terrible results since it's easy to produce abnormal latent code that mismatch the distriution of the latent space of StyleGAN.} 
\label{fig:init-ablates}
\end{figure}

Moreover, the visualization example in Fig~\ref{fig:init-ablates} also supports the qualitative results that better initialization leads to better optimization results, and faster convergency.

\begin{table}[t]
\centering
\label{tab:self-improvemetent-1}
\resizebox{\linewidth}{!}{
         \begin{tabular}{@{}l@{\hspace{.2in}}c@{\hspace{.1in}}c@{\hspace{.1in}}c@{\hspace{.1in}}c@{\hspace{.1in}}c@{\hspace{.1in}}c@{}}
        \toprule
            & \multicolumn{3}{c}{CelebA-HQ} & \multicolumn{3}{c}{CACD} \\
            \cmidrule(r){2-4} \cmidrule(r){5-7} 
            & PSNR (dB) ($\uparrow$) & SSIM ($\uparrow$) & LPIPS ($\downarrow$) & PSNR (dB) ($\uparrow$) & SSIM ($\uparrow$) & LPIPS ($\downarrow$) \\
        \midrule
        iterator started from $\overline{w}$     & 31.43    & 0.79 & 0.13 & 34.14 & 0.84 & 0.08 \\
        iterator started from $\mathcal{H}$        & \textbf{33.01}    & \textbf{0.86} & \textbf{0.10} & \textbf{35.96} & \textbf{0.92} & \textbf{0.06} \\
        \bottomrule
        \end{tabular}
}
\caption{Qualitative ablation study whether the embedding network can improves the upper-bound of the iterator's performance. Following the standard setting in Image2StyleGAN~\cite{abdal2019image2stylegan}, we treat the performance at the step 5,000 as the upper-bound.}
\end{table}

\begin{table}[t]
\centering
\label{tab:init-ablates}
\resizebox{\linewidth}{!}{
         \begin{tabular}{@{}l@{\hspace{.2in}}c@{\hspace{.1in}}c@{\hspace{.1in}}c@{\hspace{.1in}}c@{\hspace{.1in}}c@{\hspace{.1in}}c@{}}
        \toprule
            & \multicolumn{3}{c}{CelebA-HQ} & \multicolumn{3}{c}{CACD} \\
            \cmidrule(r){2-4} \cmidrule(r){5-7} 
            & PSNR (dB) ($\uparrow$) & SSIM ($\uparrow$) & LPIPS ($\downarrow$) & PSNR (dB) ($\uparrow$) & SSIM ($\uparrow$) & LPIPS ($\downarrow$) \\
        \midrule
        \textit{w/o} iterator        & 29.00                & 0.71             & 0.31             & 29.36          & 0.72          & 0.30 \\
        % Ours \textit{w/o} disentengle learning     & 29.93                & 0.77          & 0.18             & 30.30             & 0.78          & 0.17 \\
        Ours                         & \textbf{31.47}    & \textbf{0.83} & \textbf{0.16} & \textbf{32.05} & \textbf{0.83} & \textbf{0.11} \\
        \bottomrule
        \end{tabular}
}
\caption{Qualitative ablation study on the effect of the iterator on the embedding network. The baseline is that without (\textit{w/o}) the iterator, we directly supervise the embedding network using the MSE and Perceptual loss between the input image $x$ and the generated image $x_e$ from the output $w_e$ of the embedding network.}
\end{table}

\begin{figure*}[t]
\centering
\includegraphics[width=\linewidth]{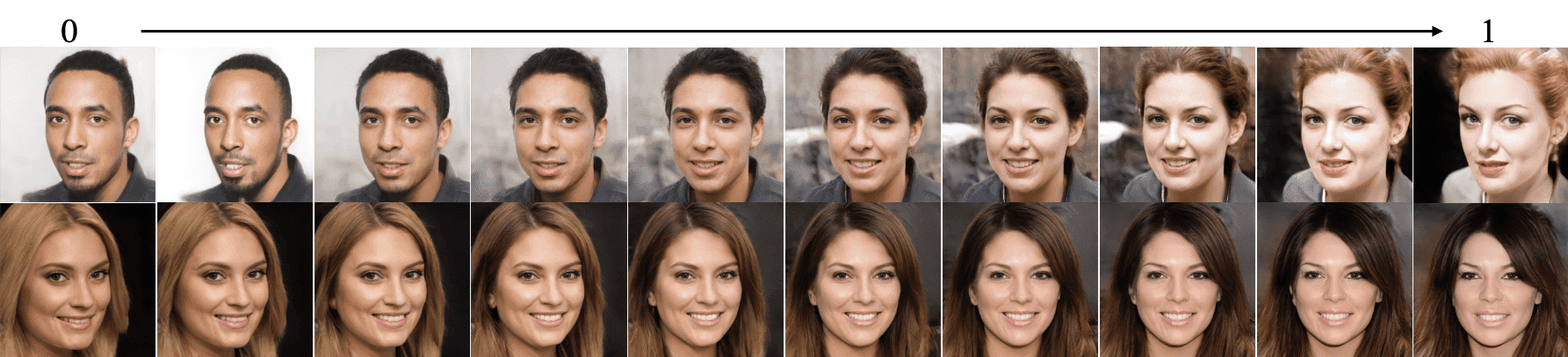}
\caption{The effect of face morphing. We gradually morph the left image to the right image by blending their latent codes. The blending parameter $\lambda$ changes from 0 to 1.} 
\label{fig:app-morphing}
\end{figure*}

\begin{figure}[t]
\centering
\includegraphics[width=0.9\linewidth]{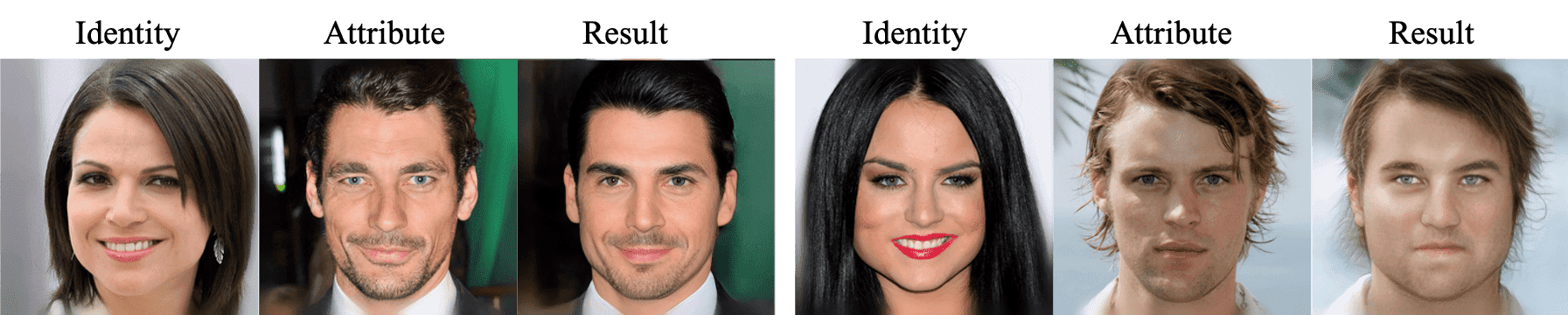}
\caption{Feature disentanglement visualization. Given one image $x_{id}$ providing ideneity and another image $x_{A}$ providing attribute, we visualize their merged results by our embedding network. From the resulst, we observe that /the merged results preserve the attribute feature (\eg hairstyle, pose, mustache \etc) from $x_{A}$ meanwhile it transfers the identity features (\eg face shape and facial features \etc) from $x_{id}$.}

% Given an image $x_{ID}$ providing identity and another image $x_{A}$ providing attribute, we use the identity encoder and the attribute encoder to extract their corresponding feature and merged them, the same as the illustrated workflow of the embedding network. From the results, we observe that /the merged results preserve the attribute feature (\eg hairstyle, pose, mustache \etc) from $x_{A}$ meanwhile it transfers the identity features (\eg face shape and facial features \etc) from $x_{ID}$. } 
\label{fig:disentengle-ablates}
\end{figure}

\begin{figure}[t]
\centering
\includegraphics[width=\linewidth]{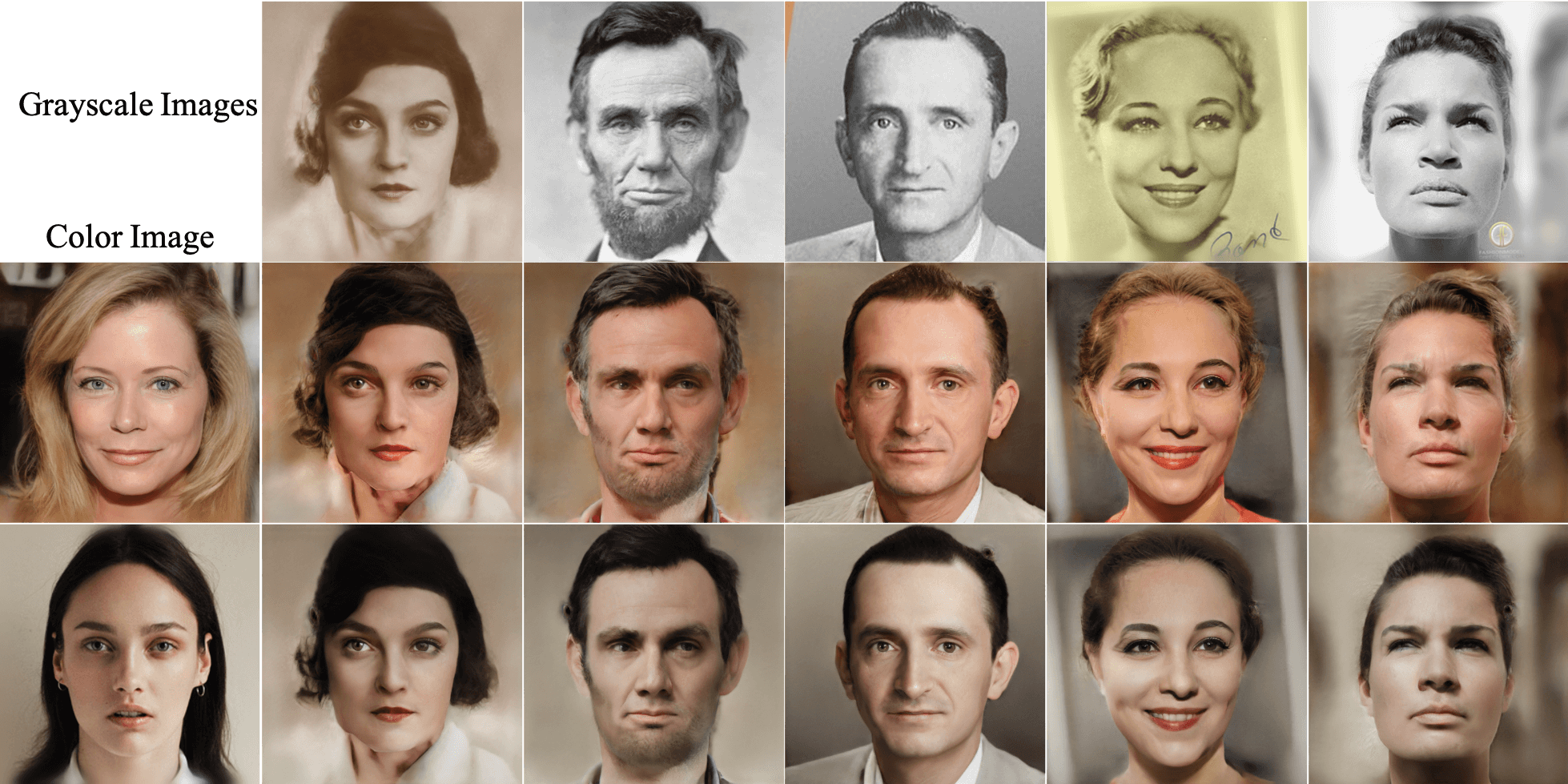}
\caption{The effect of Image colorization. The color image provides tones to colorize the grayscale images.} 
\label{fig:app-color}
\end{figure}

\noindent\paragraph{\textbf{The Evaluation on the Effect of Collaborative Learning.}}
\label{subsec:self-improve-ablates}
We conduct comparative experiments from two aspects. On the one hand, to verify whether the embedding network can improve the upbound of the iterator, we compare the iterator started from the output of out embedding network $\mathcal{H}$ (\ie, iterator started from $\mathcal{H}$) against the iterator started from a mean latent code $\bar{w}$ (\ie iterator started from $\bar{w}$). The upper-bound is termed as the performance at the step 5,000, following the setting in Image2StyleGAN. As reported in Tab.~\ref{tab:self-improvemetent-1}, the iterator started from $\mathcal{H}$ has higher upper-bound than the iterator started from $\bar{w}$. On the other hand, we ablate the iterator to examine its effect on the embedding network. The qualitative results are exhibits before in Fig.~\ref{fig:loss-gen-effect} (\ie, MSE+LPIPS). Here we report the quantitative results in Tab.~\ref{tab:init-ablates}. The embedding model without iterator takes the MSE loss and LPIPS loss between the input image and the generated image as supervision. From Tab.~\ref{tab:init-ablates}, we can observe that the performance of our full model improves a lot, thanks to the supervision on latent code provided from the iterator.

\begin{table}[t]
\centering
\label{tab:disentangle-ablates}
\resizebox{\linewidth}{!}{
         \begin{tabular}{@{}l@{\hspace{.2in}}c@{\hspace{.1in}}c@{\hspace{.1in}}c@{\hspace{.1in}}c@{\hspace{.1in}}c@{\hspace{.1in}}c@{}}
        \toprule
            & \multicolumn{3}{c}{CelebA-HQ} & \multicolumn{3}{c}{CACD} \\
            \cmidrule(r){2-4} \cmidrule(r){5-7} 
            & PSNR (dB) ($\uparrow$) & SSIM ($\uparrow$) & LPIPS ($\downarrow$) & PSNR (dB) ($\uparrow$) & SSIM ($\uparrow$) & LPIPS ($\downarrow$) \\
        \midrule
        Ours \textit{w/o} disentengle learning     & 29.93                & 0.77          & 0.20             & 30.30             & 0.78          & 0.17 \\
        Ours                 & \textbf{31.47}    & \textbf{0.83} & \textbf{0.16} & \textbf{32.05} & \textbf{0.83} & \textbf{0.11} \\
        \bottomrule
        \end{tabular}
}
\caption{Qualitative ablation study on the effect of disentangle learning. The baseline is to ablates disentangled peoduces, and directly using an Resnet encoder.}
\end{table}

\noindent \paragraph{\textbf{The Effect of Disentangled Encoders.}}
To evaluate the effect of disentanglement learning, we replace the disentangled encoder and the feature merging operation with one Resnet encoder. As shown in Tab.~\ref{tab:disentangle-ablates}, our model equipped with disentangle learning has higher performance, which mainly benefits that disentangling identity feature learning and attribute learning reduce the dimension of the learning space of encoders. Moreover, we examine the effect of disentangled feature by visualization, as shown in Fig.~\ref{fig:disentengle-ablates}. Specifically, we extract the identity feature and attribute feature from two images respectively, and then obtain their merged results using our embedding model. From the results, we can observe that the results can preserve the attribute-wise feature from attribute image, like hairstyle, pose, mustache \etc, meanwhile transfer the identity-wise features, like face shape and facial features \etc.

\section{Applications}

Previous works have demonstrated that the latent code of StyleGAN has explicitly disentangled semantic features, which is feasible for semantic image editing. InterfaceGAN has explored the attribute manipulation application based on editing the latent coder of StyleGAN. Here we additionally introduce other interesting applications below to further prove the significance of the real-time embedding network.

\noindent \paragraph{\textbf{Image Colorization.}} As shown in Fig.~\ref{fig:app-color}, given a color image providing tones, we can naturally colorize the grayscale images through a style-mixing operation. Specifically, we replace the last 10 layers of the latent codes corresponding to the grayscale image with the last 10 layers of the latent codes of the color image.

\noindent \paragraph{\textbf{Face Morphing.}} Face morphing is very valuable in video social field. Here we propose a light approach to realize morphing effect realtime in $1024 \times 1024$ resolution, shown in Fig.~\ref{fig:app-morphing}. Specifically, given two images, we use our embedding model to extract their latent codes respectively, denoted as $w_1$ and $w_2$. Then we compute the intermidiate latent code $w$ follow the function: $w = \lambda w_1 + (1-\lambda) w_2$.

\section{Conclusion}
In this work, we propose a collaborative learning framework with a carefully feature disentangle structure to learning an efficient embedding network in an unsupervised case, which enables real-time StyleGAN-based semantic image editing applications. Extensive experiments indicate that our embedding network is much faster than previous SOTA approaches while having a competitive performance with them.

{\small
\bibliographystyle{ieee_fullname}
\bibliography{egbib}
}

\end{document}